\pdfoutput=1 

\ifdefined\REVIEW
    \documentclass[
        a4paper,
    ]{article}
    \usepackage{lineno, setspace}
    \modulolinenumbers[1]
    \pagewiselinenumbers
\else
    \documentclass[
        a4paper,
        twocolumn
    ]{article}
    \usepackage{setspace}
\fi

\usepackage{IEK10} \usepackage{natbib}

\def\IEK10{
  Forschungszentrum Jülich GmbH,
  Institute of Climate and Energy Systems,
  Energy Systems Engineering (ICE-1),
  Jülich 52425,
  Germany
}
\def\SVT{
  RWTH Aachen University,
  Process Systems Engineering (AVT.SVT),
  Aachen 52074,
  Germany
}
\def\JARA{
  JARA-ENERGY,
  Jülich 52425,
  Germany
}
\def\RWTH{
  RWTH Aachen University,
  Aachen 52062,
  Germany
}

\newcommand{\mytitle}{
End-to-End Reinforcement Learning of Koopman Models for eNMPC of an Air Separation Unit
}

\newcommand{\affil}{
  \begin{itemize}[leftmargin=3mm, itemsep=0mm]
    \item[$^a$]\IEK10
    \item[$^b$]\SVT
    \item[$^c$]\JARA
    \item[$^d$]\RWTH
  \end{itemize}
}

\def\firstAuthor{Daniel Mayfrank}
\newcommand{\myauthor}{
\firstAuthor$^{a,d}$\orcidlink{0009-0000-6275-0614},
Kayra Dernek$^{a,b}$,
Laura Lang$^{b}$,
Alexander Mitsos$^{c,a,b}$\orcidlink{0000-0003-0335-6566}, 
Manuel Dahmen$^{a,*}$\orcidlink{0000-0003-2757-5253} }
\newcommand{\correspondingauthor}{Manuel Dahmen, \IEK10 \newline E-mail: m.dahmen@fz-juelich.de}
\author{\myauthor}

\usepackage[
  colorlinks,
  linkcolor=blue,
  citecolor=blue,
  urlcolor=blue,
  pdftitle={\mytitle},
  pdfauthor={\firstAuthor}
]{hyperref}
\usepackage[capitalise, nameinlink]{cleveref}
\crefname{table}{Tab.}{Tab.}

\usepackage{orcidlink}

\newcommand{\mycomment}[1]{}

\newcommand{\setpgfexternalcounter}[1]{
  \makeatletter \pgfkeysgetvalue{/tikz/external/figure name}\myexternalname
  \expandafter\gdef\csname c@tikzext@no@\myexternalname\endcsname{#1}\makeatother
}

\begin{document}

\ifx\REVIEW\undefined
\twocolumn[
\begin{@twocolumnfalse}
\fi
  \thispagestyle{firststyle}

  \begin{center}
    \begin{large}
      \textbf{\mytitle}
    \end{large} \\
    \myauthor
  \end{center}

  \vspace{0.5cm}

  \begin{footnotesize}
    \affil
  \end{footnotesize}

  \vspace{0.5cm}
  
    \textbf{Abstract} -- With our recently proposed method based on reinforcement learning (Mayfrank et al. (2024), Comput. Chem. Eng. 190), Koopman surrogate models can be trained for optimal performance in specific (economic) nonlinear model predictive control ((e)NMPC) applications. So far, our method has exclusively been demonstrated on a small-scale case study. Herein, we show that our method scales well to a more challenging demand response case study built on a large-scale model of a single-product (nitrogen) air separation unit. Across all numerical experiments, we assume observability of only a few realistically measurable plant variables. Compared to a purely system identification-based Koopman eNMPC, which generates small economic savings but frequently violates constraints, our method delivers similar economic performance while avoiding constraint violations.

  \vspace*{5mm}
\ifx\REVIEW\undefined
\end{@twocolumnfalse}
]
\fi


\section{Introduction}\label{sec:intro}
    Data-driven dynamic models can be trained in an end-to-end fashion for optimal performance as part of (economic) (nonlinear) model predictive control ((e)(N)MPC) (e.g., \cite{gros2019data, amos2018differentiable}). We recently introduced a method (\cite{mayfrank2024end}) based on reinforcement learning (RL) for end-to-end learning of Koopman \textit{surrogate} models (\cite{korda2018linear}) for (e)NMPC applications. Such data-driven surrogate models can make eNMPC computationally tractable in case an accurate mechanistic process model is available but using it as part of a model predictive controller is too computationally expensive. Moreover, in scenarios where no reliable model is available, the same framework can learn directly from plant data. Alternative methods for end-to-end learning of data-driven models for control applications focus on linear models (\cite{chen2019gnu}), optimize highly-structured models with few parameters requiring expert system knowledge (\cite{brandner2023reinforcement}), cannot handle hard constraints on system states (\cite{amos2018differentiable}), or are only applicable to setpoint tracking problems (\cite{iwata2022data}). Our method can optimize highly parameterized and thus flexible Koopman models for control problems with state constraints and arbitrary convex objective functions.

    In \cite{mayfrank2024end}, we demonstrated our method in two simulated case studies (NMPC and eNMPC) based on a small model of a continuous stirred-tank reactor (CSTR) (\cite{flores2006simultaneous}) comprised of just two ordinary differential equations. The resulting policies outperformed Koopman controllers employing models that were trained using the prevailing system identification (SI) approach by (i) achieving more accurate state tracking in the NMPC case study and (ii) substantially reducing the frequency of constraint violations in the eNMPC case study. In the present contribution, we demonstrate the scalability of our method (\cite{mayfrank2024end}) using a large-scale differential-algebraic equations (DAE) model of an air separation unit (ASU) (\cite{CASPARITopDown}).

    The remainder of this short paper is organized as follows: First, the ASU demand response case study is introduced in Sec. \ref{sec:casestudy}. Then, Sec. \ref{sec:method} provides a brief explanation of our method, followed by a description of the adjustments to the Koopman model architecture we employed in this work. Sec. \ref{sec:experiments} presents the results of the numerical experiments. Finally, Sec. \ref{sec:conclusion} discusses the conclusions and directions for future work.

\section{Demand response of an air separation unit}\label{sec:casestudy}
    We consider demand response of a single-product ASU for the production of purified nitrogen based on the benchmark process presented in \cite{CASPARITopDown}, resulting in a mid-level complexity control problem. The process flowsheet is shown in Figure \ref{fig:ASU}. Because RL approaches often need many policy-environment interactions to produce good results, wall-clock simulation speed is critical for simulation models that are to be used as part of the training environment. Because the full mechanistic model \cite{CASPARITopDown} is computationally expensive, we construct the demand-response RL environment using the modified model of \cite{schulze2023data}, which enables substantially faster simulation. The modified model is a nonlinear DAE system with 2327 algebraic and 119 differential states, implemented in Modelica and still computationally expensive, thus motivating the proposed end-to-end learning.

    \mycomment{
    \begin{figure*}[htb]
    	\centering
    	\includegraphics[width=0.5\paperwidth]{./ASU_flowsheet.pdf}
    	\caption{Air separation process flowsheet. The following manipulated variables are shown with grey background: inlet air flow rate $F_\text{mac}$, air fraction passing through turbine 1 $\xi_\text{phx}$, distillation column reflux ratio $\xi_\text{cond}$, drain stream $F_\text{dr}$. The controlled variables are depicted with dotted circles: product impurity $I_\text{prod}$, molar holdup in storage $N_\text{s}$ and reboiler $N_\text{r}$, temperature difference between reboiler and condenser $\Delta T_{\text{rc}}$.}
    	\label{fig:ASU}
    \end{figure*}
    }

    \begin{figure}[htb]
    	\centering
    	\includegraphics[width=0.39\paperwidth]{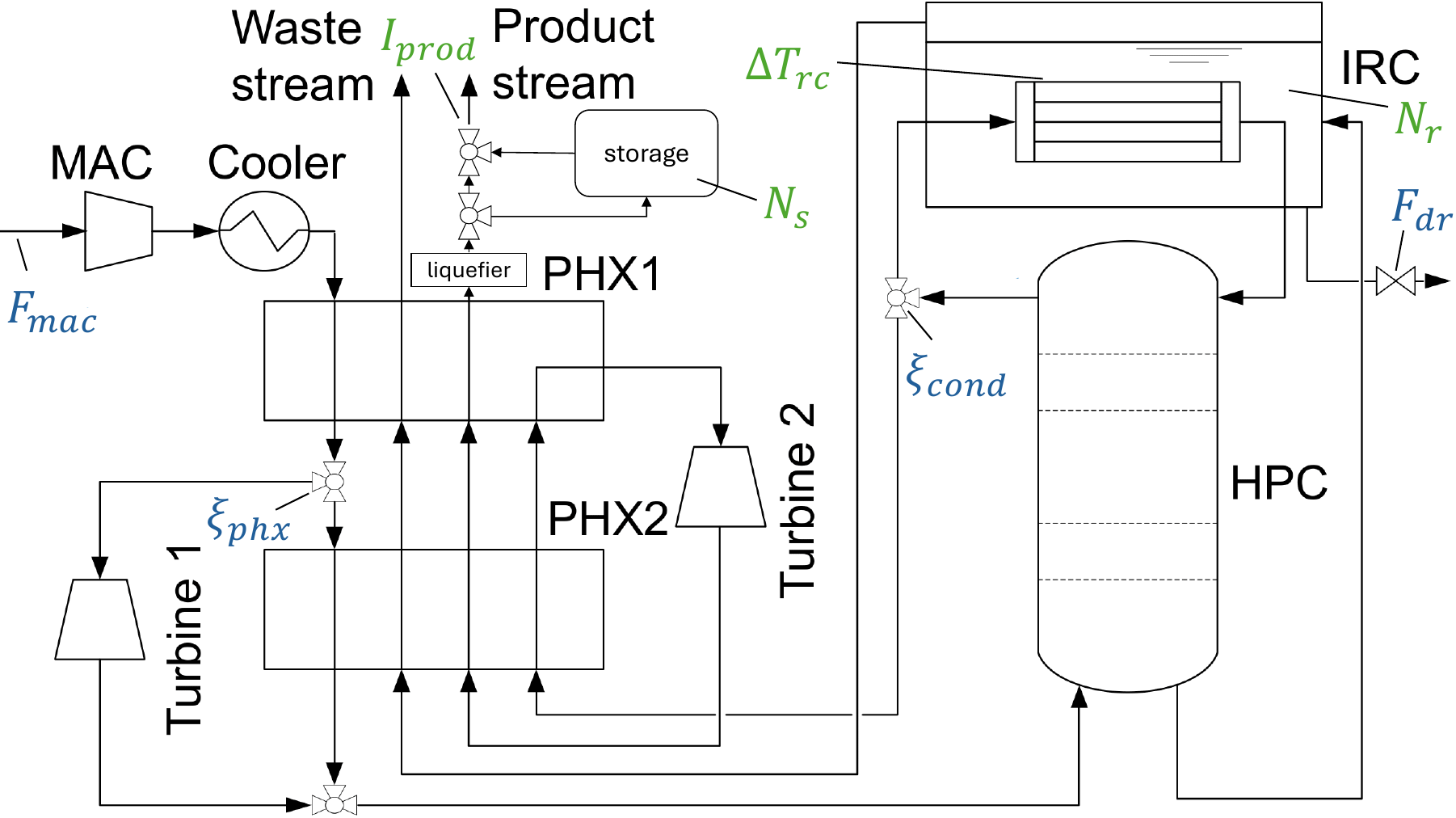}
    	\caption{Air separation process flowsheet. The following manipulated variables are shown in \textcolor{blue}{blue} font: inlet air flow rate $F_\text{mac}$, air fraction passing through turbine 1 $\xi_\text{phx}$, distillation column reflux ratio $\xi_\text{cond}$, drain stream $F_\text{dr}$. The controlled variables are depicted in \textcolor{green}{green} font: product impurity $I_\text{prod}$, molar holdup in storage $N_\text{s}$ and reboiler $N_\text{r}$, temperature difference between reboiler and condenser $\Delta T_{\text{rc}}$.}
    	\label{fig:ASU}
    \end{figure}

    Ambient air is compressed in the main air compressor (MAC), pre-cooled, and then passes through a two-part multi-stream heat exchanger (MSHE), where it is cooled against returning process streams. After the first part of the MSHE (PHX1), a fraction of the air is used in turbine 1 for power generation, while the remainder is liquefied in the second part of the MSHE (PHX2). Both streams are recombined before entering the high-pressure distillation column (HPC). The oxygen-rich bottom product of the HPC is expanded and used in an integrated reboiler condenser (IRC) to cool the reflux stream. Liquid is withdrawn via a drain stream, and the exiting vapor leaves the process as waste after heat recovery in the MSHE. The nitrogen-rich top product passes through turbine 2 and a liquefier to yield liquid nitrogen, which can be stored in a product tank for flexible delivery.

    The task of the eNMPC is to minimize operational cost by exploiting variations in the electricity price, while fulfilling a constant demand for liquid nitrogen and avoiding constraint violations. The operational cost is given by the overall power consumption $E$ of the ASU multiplied by the electricity price. For the overall power consumption, the energy demand of the MAC and the liquefier, as well as the electricity generation from the turbines, are taken into account. 
    Operational constraints and manipulated control inputs are shown in the ASU flowsheet in Fig. \ref{fig:ASU}, and their respective lower and upper bounds are given in Table \ref{tab:operational_constraints}.

    \begin{table}[h]
        \centering
        \resizebox{0.33\paperwidth}{!}{
        \begin{tabular}{llll}
            \toprule
            Variable & lb & ub & Constraint type \\
            \midrule
            $I_{\text{prod}}$ [ppm]      & 0                   & 1800                  & path     \\
            $\Delta T_{\text{rc}}$ [K]     & 2                 & 5                     & path     \\
            $N_r$ [kmol]                     & 2               & 10                    & path     \\
            $N_s$ [-]                       & 0                & 6                     & path     \\
            $F_{\text{mac}}$ [mol/s]      & 30                   & 50                  & input     \\
            $F_{\text{dr}}$ [mol/s]       & 0                    & 2                   & input     \\
            $\xi_{\text{phx}}$ [kmol]    & 0                   & 0.1                  & input     \\
            $\xi_{\text{cond}}$ [-]       & 0.51                & 0.54                 & input     \\
            \bottomrule
        \end{tabular}
        }
        \caption{Summary of lower (lb) and upper (ub) bounds of the operational and input variables.}
        \label{tab:operational_constraints}
    \end{table}

    To use the Modelica model as part of an RL environment, we export it as a functional mock-up unit that can be simulated within Python code. At each control step in the environment, the policy receives the current state of the ASU and an electricity price prediction for the upcoming 9 hours in hourly resolution. After receiving a control input from the policy, the state of the ASU is updated by simulating the model for a time step of $\SI{15}{min}$. Furthermore, analogous to the reward calculation in \cite{mayfrank2024end}, we calculate a reward based on constraint violations and electricity cost savings compared to steady-state production.

\section{End-to-end RL of Koopman Model for eNMPC}\label{sec:method}
    In \cite{mayfrank2024end}, we utilize Koopman models of the form proposed by \cite{korda2018linear}: (i) A nonlinear state observation function $\bm{\psi}_{\bm{\theta}}\colon \mathbb{R}^{n_{\bm{x}}} \mapsto \mathbb{R}^{n_{\bm{z}}}$ that transforms the initial system state $\bm{x}_0 \in \mathbb{R}^{n_{\bm{x}}}$ into the initial Koopman state $\bm{z}_0 \in \mathbb{R}^{n_{\bm{z}}}$, where typically $n_{\bm{z}} \gg n_{\bm{x}}$: $\bm{z}_0 = \bm{\psi}_{\bm{\theta}}(\bm{x}_0)$. (ii) The $\bm{A}_{\bm{\theta}} \in \mathbb{R}^{n_{\bm{z}} \times n_{\bm{z}}}$ and $\bm{B}_{\bm{\theta}} \in \mathbb{R}^{n_{\bm{z}} \times n_{\bm{u}}}$ matrices, which linearly approximate the evolution of the Koopman state, driven by external control inputs $\bm{u}_t\in \mathbb{R}^{n_{\bm{u}}}$: $\bm{z}_{t+1} = \bm{A}_{\bm{\theta}} \bm{z}_t + \bm{B}_{\bm{\theta}} \bm{u}_t$. (iii) The $\bm{C}_{\bm{\theta}} \in \mathbb{R}^{n_{\bm{x}} \times n_{\bm{z}}}$ matrix, which linearly transforms the Koopman state $\bm{z}_t$ into a predicted system state $\hat{\bm{x}}_t$: $\hat{\bm{x}}_t = \bm{C}_{\bm{\theta}} \bm{z}_t$. Such models can be trained by adjusting the parameters $\bm{\theta}$.

    RL algorithms (e.g., \cite{schulman2017proximal}) can be used to optimize parameterized policies $\bm{\pi}_{\bm{\theta}}(\bm{u}_{t}|\bm{x}_{t})\colon \mathbb{R}^{n_{\bm{x}}} \mapsto \mathbb{R}^{n_{\bm{u}}}$, which map current system states ${\bm x}_t$ to control actions ${\bm u}_t$, by maximizing the expected cumulative reward of the policy. Therefore, to train a Koopman model in an end-to-end manner for a particular eNMPC application via RL, we construct an (automatically differentiable) eNMPC policy based on the Koopman model. When used as part of an eNMPC policy, Koopman models of the above stated form (\cite{korda2018linear}) result in convex optimal control problems (OCPs) if the stage cost and all additional user-defined constraints are convex. By using the automatically differentiable solver \textit{cvxpylayers} (\cite{Agrawal2019differentiable}) for convex optimization problems, we can obtain $\partial \bm{u}^*_t /\partial \bm{\theta}$ via implicit differentiation of the KKT conditions. The solution map $\bm{\theta}\mapsto \bm{u}^*_t$ is only piecewise differentiable; nevertheless our numerical experience shows that the values returned from \textit{cvxpylayers} as derivatives work well for the first-order training.

    In our previous work \citep{mayfrank2024end}, the path constraints for the controlled variables were formulated as inequality constraints. However, when applied to the ASU model, the end-to-end learning approach showed no improvement over the initial guess. We attribute this to poor gradient estimation caused by switching between active and inactive inequality constraints. To address this issue, we reformulate the OCP in the end-to-end RL by replacing the inequality constraints with equality constraints with slack variables $\bm{s}$, and penalize the slack variables through quadratically smoothed hinge loss (\cite{Zhang_2004}), defined as \(     L(s_i, \delta) = M \left[\max\left(0,\, |s_i| - \frac{1}{2} \Delta g_i + \delta\right)\right]^2 \), where $\Delta g_i$ represents the admissible range of the corresponding constraint (see supplementary material). The variable \( M > 0 \) is a penalty coefficient to balance the penalty and the objective performance. To more strongly penalize constraint violations, we introduce penalty scaling values $\delta$ that slightly tighten the bounds of the inequality constraints, leading to a more conservative control behavior.

\begin{figure}[htb]
    	\centering
    	\includegraphics[width=0.26\paperwidth]{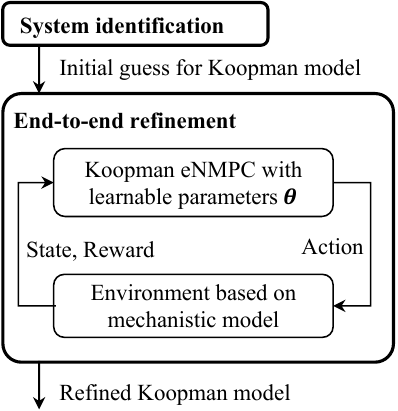}
    	\caption{Workflow for end-to-end learning of a Koopman model for eNMPC.}
    	\label{fig:workflow}
    \end{figure}

    Our workflow is shown in Fig. \ref{fig:workflow}.
    Initial values for $\bm{\theta}$ are obtained through standard system identification (SI), i.e., we generate simulation data using the mechanistic model and fit $\bm{\theta}$ to that data. A more detailed description of the SI is provided in the supplementary materials. The model parameters $\bm{\theta}$ are then fine-tuned for optimal performance in a specific control task using Proximal Policy Optimization (PPO) \citep{schulman2017proximal}, an actor-critic RL algorithm.

\begin{figure}[htb]
    	\centering
    	\includegraphics[width=0.26\paperwidth]{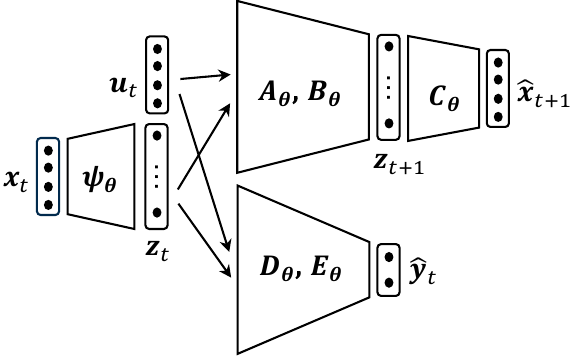}
    	\caption{Koopman model architecture.}
    	\label{fig:Koopman_architecture}
    \end{figure}

    The ASU model is a DAE system that features system outputs $\bm{y}_t$ that exhibit discontinuities when control inputs change non-continuously. The Koopman model architecture proposed by \cite{korda2018linear}, which we used in \cite{mayfrank2024end}, cannot reflect such behavior since the control inputs enter directly into the predictor-part of the model, which produces the latent space vector one time step into the future. To enable the Koopman models to represent instantaneous output responses to input changes, we extend the framework of \cite{korda2018linear} by adding a second decoder, i.e., $\bm{y}_t = \bm{D}_{\bm{\theta}} \bm{z}_t + \bm{E}_{\bm{\theta}} \bm{u}_t$, with learnable $\bm{D}_{\bm{\theta}} \in \mathbb{R}^{n_{\bm{y}} \times n_{\bm{z}}}$ and $\bm{E}_{\bm{\theta}} \in \mathbb{R}^{n_{\bm{y}} \times n_{\bm{u}}}$ matrices. Note that this second decoder was not needed in our previous work \citep{mayfrank2024end} because the CSTR investigated there was described by an ordinary differential equation (ODE) system for which discontinuities in control inputs do not cause discontinuities in states/outputs. The overall architecture of the Koopman model is visualized in Figure \ref{fig:Koopman_architecture}.

    We do not assume full observability of the ASU. Instead, we use only the following states as inputs to the Koopman surrogate model: the product purity $I_{\text{prod}}$, the temperature difference between reboiler and condenser $\Delta T_{\text{rc}}$, the reboiler holdup $N_{\text{r}}$, and the temperature of tray 20 in the distillation column $T_{\text{tray,20}}$. Based on this information, the model constructs a latent representation $\bm{z}_t$ of the state of the ASU. The vector of control inputs $\bm{u}_t$ consists of the four inputs listed in Table \ref{tab:operational_constraints}. The model predicts the evolution of the control-relevant entries of $\bm{x}_t$ through the $\bm{A}_{\bm{\theta}}$, $\bm{B}_{\bm{\theta}}$, and $\bm{C}_{\bm{\theta}}$ matrices. The resulting vector $\hat{\bm{x}}_t \in \mathbb{R}^{3}$ consists of $I_{\text{prod}}$, $\Delta T_{\text{rc}}$, and $N_r$. The control-relevant, discontinuous outputs $\bm{y}_t \in \mathbb{R}^{2}$ are the energy demand $E$ of the process and the production rate $\dot{n}_{product}$. The latter is used to calculate the molar holdup $N_\text{s}$ of the storage tank through a mass balance, where the change in tank storage is given by the difference between the inflow and outflow rates, i.e., \(
        \frac{dN_\text{s}}{dt} = \dot{n}_{\text{product}} - \dot{n}_{\text{demand}}.\)
    The molar holdup $N_\text{s}$ is upper and lower bounded to reflect the physical storage capacity limits of the tank.

    We choose a latent space dimensionality of 10 for the Koopman model, i.e., $\bm{z}_t \in \mathbb{R}^{10}$. Altogether, the Koopman model thus consists of the matrices $\bm{A}_{\bm{\theta}} \in \mathbb{R}^{10 \times 10}, \bm{B}_{\bm{\theta}} \in \mathbb{R}^{10 \times 4}, \bm{C}_{\bm{\theta}} \in \mathbb{R}^{3 \times 10}, \bm{D}_{\bm{\theta}} \in \mathbb{R}^{2 \times 10}, \bm{E}_{\bm{\theta}} \in \mathbb{R}^{2 \times 4}$, and an encoder $\bm{\psi_\theta}\colon \mathbb{R}^{4} \to \mathbb{R}^{10}$. The encoder is a feedforward neural network with two hidden layers with 50 neurons each, and tanh activation functions.

\section{Numerical experiments}\label{sec:experiments}
    All training code used in this work, including the implementation of the ASU demand response RL environment, is available online\footnote{\url{https://jugit.fz-juelich.de/iek-10/public/optimization/koopmanenmpc4asu}}.

    We obtain an initial guess for the Koopman model via the iterative data sampling and system identification approach outlined in the supporting information. Then, we repeat the end-to-end refinement five times using different fixed seeds in every training run. We use the historic German day-ahead electricity prices from 2023-01-01 to 2023-12-31 (\cite{epex2023}) for training.

\begin{figure}[htb]
    	\centering
    	\includegraphics[width=0.32\paperwidth]{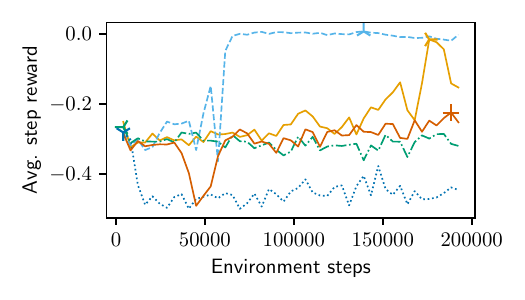}
    	\caption{Learning progress during end-to-end model refinement. Each line represents one training run. For each training run, a marker depicts the highest average reward achieved in one policy rollout.}
    	\label{fig:training_rewards}
    \end{figure}

    Fig. \ref{fig:training_rewards} illustrates the evolution of the reward over 200,000 environment steps for each training run. Each training run takes approx. two days on a Windows 11 workstation with an Intel Core i7-14700 CPU. Three elements contribute to overall training time: (i) simulation of the mechanistic DAE environment, (ii) repeated solution of the convex eNMPC problem, and (iii) gradient-based policy optimization. The dominant contributor to the training time is the policy optimization step, which requires backpropagating through the Koopman eNMPC. In contrast, pure policy inference typically takes well under one second per control step, making the controller easily real-time capable. After training has completed, our final performance evaluation is based on the policy that attained the highest average reward during a policy rollout. Consequently, our primary metric of interest is the maximum average reward achieved by each training run, rather than the reward at the final training step. Three out of five training runs demonstrate a substantial improvement in the policy’s average reward compared to the initial guess.
    After training, we test the performance of the policy that obtained the highest average reward in a three-day test episode. The test electricity price trajectory is generated using the method by \cite{papadimitriou2024representative}, which constructs representative price profiles from 2023 historical data while preserving key statistics (mean and variance). The resulting profiles are distinct from those used during training, ensuring a clear separation between training and testing data.

    In the test episode, the eNMPC employing the Koopman model obtained solely via system identification (from hereon called \textit{Koopman-SI} policy) saves $\SI{1}{\%}$ of electricity cost compared to steady-state production while producing small constraint violations in $\SI{16.3}{\%}$ of the time steps, resulting in an average reward of $-0.26$. The eNMPC employing the end-to-end refined Koopman model (\textit{Koopman-PPO}) improves performance, and produces $\SI{2}{\%}$ cost savings. Furthermore, it does so without violating any constraints in the test episode, thus yielding an average reward of $0.01$.

\begin{figure}[htb]
    	\centering
    	\includegraphics[width=0.32\paperwidth]{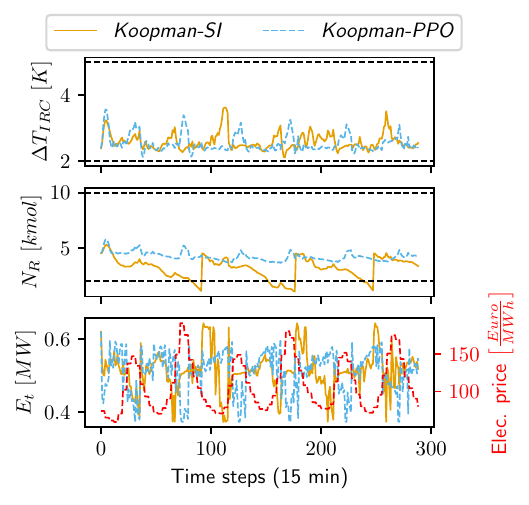}
    	\caption{Comparison of the control behavior of \textit{Koopman-SI} and \textit{Koopman-PPO}.}
    	\label{fig:trajectories}
    \end{figure}

    Figure \ref{fig:trajectories} illustrates the behavior of the policies in the three-day test episode. We show the evolution of $\Delta T_{IRC}$ and $N_R$, since these are the only states that operate close to their bounds, as well as the evolution of the energy demand $E$. Both policies exhibit an intuitive inverse relationship between the electricity price and the energy demand. It can be seen that the behavior of the policies differs for states $\Delta T_{IRC}$ and $N_R$. While \textit{Koopman-SI} exhibits violations in $N_R$, \textit{Koopman-PPO} maintains operation safely within the operation bounds. Both policies exhibit high-frequency oscillations with respect to $E_t$.
    
    As stated above, the architecture of the Koopman models that we use (see Figure \ref{fig:Koopman_architecture}) results in convex OCPs, which are relatively easy to solve. In the 72 hour test episode, i.e., 288 control steps of $\SI{15}{min}$ length each, the inference time of the \textit{Koopman-PPO} policy was between $\SI{0.1}{s}$ and $\SI{3}{s}$ with an average time of $\SI{0.5}{s}$.

\section{Conclusion}\label{sec:conclusion}
    We apply our previously published method (\cite{mayfrank2024end}) for end-to-end learning of Koopman surrogate models for (e)NMPC applications to a high-dimensional nonlinear demand response problem based on a mechanistic model of an ASU, demonstrating its applicability to complex, real-world systems.

    Among the five independent end-to-end training runs, three significantly improve eNMPC performance over the SI baseline, reducing and even eliminating constraint violations while maintaining high economic efficiency. Compared to our earlier two-state CSTR case study (\cite{mayfrank2024end}), where all ten training runs consistently improved performance and rewards saturated after roughly 180,000 steps, the present large-scale ASU problem poses a more challenging learning task, reflected in the lower fraction of successful runs. Note that while our method strongly incentivizes constraint satisfaction, it does \textit{not} provide formal theoretical guarantees.

    Our method produces data-driven predictive control policies that strike an exceptional balance between control performance (high economic performance and consistent constraint satisfaction) and computational efficiency (policy inference time $\ll$ sampling time). By demonstrating its scalability to an industrially-relevant process, we show that our method could be applicable to complex real-world control problems where mechanistic predictive control policies are not real-time capable and system identification-based data-driven controllers produce unsatisfactory performance.

    Future work should test our method on a real-world process. Because the Koopman surrogate model is trained in an environment based on a mechanistic simulation model, it may overfit to systematic simulation errors, e.g., parameter mismatch or unmodeled dynamics. The resulting controller could perform suboptimally on the real process. This risk does not exist in the present work, since training and final evaluation use the same simulated environment. In the described real-world scenario, our method may be used to further refine the surrogate model via interactions with the real process. In such an approach, the sample efficiency of the learning procedure would become critical, which could motivate employing a model-based RL algorithm as done in \cite{mayfrank2025sample}.

\section*{Declaration of Competing Interest}
We have no conflict of interest.

\section*{Acknowledgements}
This work was performed as part of the Helmholtz School for Data Science in Life, Earth and Energy (HDS-LEE) and received funding from the Helmholtz Association of German Research Centres. Additionally, the authors gratefully acknowledge the financial support of the Kopernikus project SynErgie by the Federal Ministry of Education and Research (BMBF), and the project supervision by the project management organization Projektträger Jülich (PtJ).


\section*{Author contributions}
Daniel Mayfrank: Conceptualization, Methodology, Software, Investigation, Writing - original draft,  Writing - review \& editing, Visualization

\noindent Kayra Dernek: Methodology, Software, Investigation, Writing - review \& editing, Visualization

\noindent Laura Lang: Methodology, Writing - review \& editing, Writing - original draft

\noindent Alexander Mitsos: Conceptualization, Writing - review \& editing, Supervision, Funding acquisition

\noindent Manuel Dahmen: Conceptualization, Methodology, Writing - review \& editing, Supervision, Funding acquisition

\appendix

\bibliographystyle{apalike}
  \renewcommand{\refname}{Bibliography}

  \bibliography{bibliography.bib}

\end{document}


\ifx\REVIEW\undefined
\twocolumn[
\begin{@twocolumnfalse}
\fi
  \thispagestyle{firststyle}

  \begin{center}
    \begin{large}
      \textbf{\mytitle}
    \end{large} \\
    \myauthor
  \end{center}

  \vspace{0.5cm}

  \begin{footnotesize}
    \affil
  \end{footnotesize}

\noindent
\textbf{Keywords:}
Economic model predictive control;
Koopman;
Demand response;
Air separation unit
  \vspace*{5mm}
\ifx\REVIEW\undefined
\end{@twocolumnfalse}
]
\fi


\section{System identification procedure}
\begin{figure}[htb]
    	\centering
    	\includegraphics[width=0.32\paperwidth]{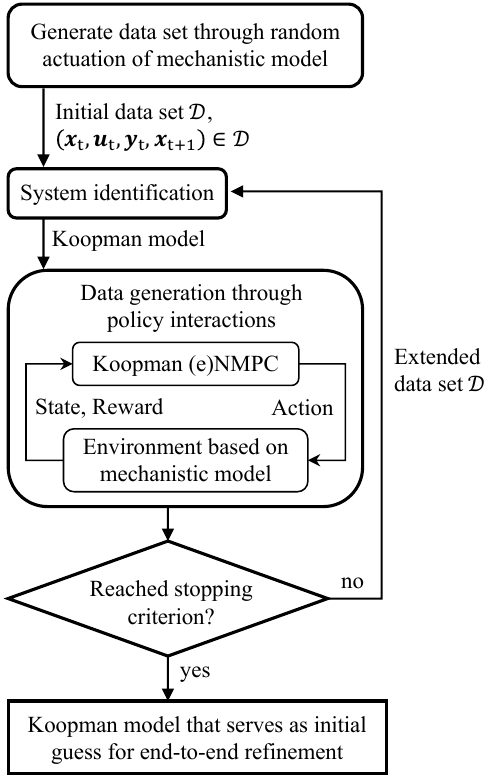}
    	\caption{Iterative data sampling and system identification procedure.}
    	\label{fig:SI_workflow}
    \end{figure}

    In this work, we employ the iterative data sampling and system identification approach depicted in Figure \ref{fig:SI_workflow}: We start by generating data through random actuation of the mechanistic model, followed by fitting the parameters $\bm{\theta}$ of the Koopman model to the obtained data. Then, we construct an eNMPC policy based on the Koopman model and let the policy interact with the environment for 2880 time steps, i.e., 30 simulated days, to extend the training data set, and we retrain the Koopman model on this larger data set. We keep extending the data set and retraining our Koopman model with the extended data set until the maximum average reward obtained during data sampling in one iteration does not improve for five consecutive iterations. The model of the policy that produced the maximum average reward is then used as the initial guess for the RL-based end-to-end learning procedure. At this stage, we do not employ the inequality constraints reformulation presented later; instead, we follow the same optimal control problem (OCP) formulation as in our previous work \citep{mayfrank2024end}.
\section{Chaining of model predictions}
    RL-based end-to-end refinement of the Koopman model requires solving and backpropagating through the OCPs numerous times. To decrease the associated computational burden, it makes sense to minimize the number of optimization variables in the OCP, i.e., to maximize the time step duration $\Delta t$ of the Koopman model and the resulting eNMPC controller. We observe that a discretization of the eNMPC controller using $\SI{15}{\text{min}}$ time steps is short enough to enable the controller to make use of the full feasible space of control inputs without causing an excessive number of constraint violations due to too fast process dynamics. However, during preliminary testing of system identification, we observe that we obtain more accurate model predictions if we chain the predictions of models that predict shorter time steps. Therefore, we take the following approach: During system identification, we generate data with a $\SI{5}{\text{min}}$ discretization and we fit a Koopman model to that data. However, before using the model as part of eNMPC (in data sampling or end-to-end refinement), we transform the prediction components of the Koopman model ($\bm{A}, \bm{B}, \bm{D}, \bm{E}$) to predict $\SI{15}{\text{min}}$ time steps instead of $\SI{5}{\text{min}}$ steps, i.e., a one-step prediction of the transformed model is equivalent to chaining three predictions of the original model with constant control input. The following equations are used for this exact transformation:
    \begin{subequations}\label{eq:model_upscaling}
    \begin{align}
        \bm{A}_{\text{15min}} &= \bm{A}^3, \\
        \bm{B}_{\text{15min}} &= \bm{A}^2 \bm{B} + \bm{A} \bm{B} + \bm{B}, \\
        \bm{D}_{\text{15min}} &= \bm{D} \bm{A}^3, \\
        \bm{E}_{\text{15min}} &= \bm{D} \bm{A}^2 \bm{B} + \bm{D A B} + \bm{D B} + \bm{E}.
    \end{align}
    \end{subequations}
    Eqs. \ref{eq:model_upscaling} are derived in Eqs. \ref{eq:model_upscaling_derivation_AB} and Eqs. \ref{eq:model_upscaling_derivation_DE}. Please note that the eNMPC has control steps of $\SI{15}{\text{min}}$ length, and therefore, $\bm{u}_{t} = \bm{u}_{t+1} = \bm{u}_{t+2}$ in the $\SI{5}{\text{min}}$ discretization Koopman model. To perform the upscaling of the Koopman model from a $\SI{5}{\text{min}}$ time step to a $\SI{15}{\text{min}}$ time step, we derive the transformed system dynamics by chaining the predictions of the original model over three consecutive time steps with constant input. The following equations describe the stepwise progression for the latent state $\bm{z}_t$ and output $\bm{y}_t$ variables over three time steps, with constant control input $\bm{u}_t$ over the interval. This allows us to predict the system’s behavior at the 15-minute time scale while maintaining consistency with the original model dynamics.
    \begin{subequations}\label{eq:model_upscaling_derivation_AB}
    \begin{align} \label{eqn:cubed_matrices}
        \bm{z}_{t+1} &= \bm{A} \bm{z}_t + \bm{B} \bm{u}_t \\
        \bm{z}_{t+2} &= \bm{A} (\bm{A} \bm{z}_t + \bm{B} \bm{u}_t) + \bm{B} \bm{u}_t = \bm{A}^2 \bm{z}_t + \bm{A} \bm{B} \bm{u}_t + \bm{B} \bm{u}_t \\
        \begin{split}
            \bm{z}_{t+3} &= \bm{A} (\bm{A}^2 \bm{z}_t + \bm{A B} \bm{u}_t + \bm{B} \bm{u}_t) + \bm{B} \bm{u}_t = \bm{A}^3 \bm{z}_t + \bm{A}^2 \bm{B} \bm{u}_t + \bm{A B} \bm{u}_t + \bm{B} \bm{u}_t \\
            &= \bm{A}^3 \bm{z}_t + (\bm{A}^2 \bm{B} + \bm{A} \bm{B} + \bm{B}) \bm{u}_t = \bm{A}_{\text{15min}} \bm{z}_t + \bm{B}_{\text{15min}} \bm{u}_t
        \end{split}\\
        &\Rightarrow \bm{A}_{\text{15min}} = \bm{A}^3, \bm{B}_{\text{15min}} = \bm{A}^2 \bm{B} + \bm{A} \bm{B} + \bm{B}
    \end{align}
    \end{subequations}

    \begin{subequations}\label{eq:model_upscaling_derivation_DE}
    \begin{align}
        \bm{y}_{t+1} &= \bm{D} \bm{z}_{t+1} + \bm{E} \bm{u}_t \\
        \bm{y}_{t+2} &= \bm{D} \bm{z}_{t+2} + \bm{E} \bm{u}_t \\
        \begin{split}
            \bm{y}_{t+3} &= \bm{D} \bm{z}_{t+3} + \bm{E} \bm{u}_t = \bm{D} (\bm{A}^3 \bm{z}_t + \bm{A}^2 \bm{B} \bm{u}_t + \bm{A B} \bm{u}_t + \bm{B} \bm{u}_t) + \bm{E} \bm{u}_t \\
            &= \bm{D} \bm{A}^3 \bm{z}_t + (\bm{D} \bm{A}^2 \bm{B} + \bm{D A B} + \bm{D B} + \bm{E}) \bm{u}_t = \bm{D}_{\text{15min}} \bm{z}_t + \bm{E}_{\text{15min}} \bm{u}_t
        \end{split}\\
        &\Rightarrow \bm{D}_{\text{15min}} = \bm{D} \bm{A}^3, \bm{E}_{\text{15min}} = \bm{D} \bm{A}^2 \bm{B} + \bm{D A B} + \bm{D B} + \bm{E}
    \end{align}
    \end{subequations}

\section{Reformulation of Inequality Constraints as Equality Constraints}

    In this work, inequality constraints for the controlled variables, i.e., product impurity $I_\text{prod}$, molar holdup in storage $N_\text{s}$ and reboiler $N_\text{r}$, and temperature difference between reboiler and condenser $\Delta T_{\text{rc}}$, are converted into equality constraints through the introduction of slack variables (see main text). The reformulation is done as follows:

\noindent
Each inequality constraint of the form
\begin{equation}
    g_i^{\text{min}} \le g_i(y) \le g_i^{\text{max}},
\end{equation}
is reformulated as an equality constraint by introducing a slack variable \( s_i \):
\begin{equation}
    g_i(y) + s_i = \frac{1}{2}(g_i^{\text{min}} +  g_i^{\text{max}}).
\end{equation}
We penalize the slack variable with the quadratically smoothed hinge loss penalty function (\cite{Zhang_2004}):

\begin{equation}
    L(s_i, \delta) = M \left[\max\left(0,\, |s_i| - \frac{1}{2} (g_i^{\text{max}} - g_i^{\text{min}}) + \delta\right)\right]^2
\end{equation}
    The penalty coefficient \( M > 0 \) is set to 10{,}000 to balance the penalty term with the magnitude of the objective function, which represents the cost savings. The penalty scaling factor \( \delta \) is chosen as 0.2 to impose stronger penalties on constraint violations. Consequently, this results in effectively tighter bounds and a more conservative control behavior. In our reinforcement learning framework, all states except for the storage state, and control actions are scaled to the range \([-1, 1]\). The calculations of the slack variables and penalties are imposed on the scaled values of the controlled variables. The storage state is expressed as the rate of hourly demand, and its contribution to the penalty term is calculated in its unscaled value.
    
\section{Hyperparameters}

\begin{table}[h!]
\centering
\caption{Hyperparameters adapted from our previous work \citep{mayfrank2024end}. Where possible, the notation is consistent with the PPO paper \citep{schulman2017proximal}.}
\begin{tabular}{lll}
\hline
\textbf{Hyperparameter} & \textbf{Value} & \textbf{Description} \\
\hline
\multicolumn{3}{l}{\textbf{General}} \\
$\beta$ & $5 \cdot 10^{-5}$ & reward calculation hyperparameter \\
$\sigma$ & $(0.15, 0.15, 0.15, 0.15)^{\mathsf{T}}$ & standard deviation for action selection \\
$\gamma$ & $0.98$ & reward discount factor \\
$\lambda$ & $0.95$ & generalized advantage estimation hyperparameter \\
$\epsilon$ & $0.2$ & clipping hyperparameter \\
VF coeff. & $5.0$ & value function coefficient \\
Entropy coeff. & $10^{-3}$ & Entropy coefficient \\
$N_{\mathrm{PPO}}$ & $8$ & number of parallel actors \\
$T_{\mathrm{PPO}}$ & $512$ & control steps between updates to actor and critic \\
$M_{\mathrm{PPO}}$ & $256$ & minibatch size \\
optimizer & Adam & optimizer used for updates to actor and critic \\
$K_{\mathrm{PPO}}$ & $10$ & number of epochs per update \\
$\alpha$ & $10^{-4}$ & learning rate  \\
max. gradient norm & $0.5$ & gradient clipping value for actor update \\
Koopman MPC solver & ECOS, SCS & solver for Koopman OCPs \\
& & \citep{Domahidi2013, SCS_2016} \\
$M$ & $10{,}000$ & penalty factor for slack variable usage \\
$\delta$ & $0.2$ & penalty scaling factor\\
\hline
\end{tabular}
\end{table}

\clearpage
\appendix

\bibliographystyle{apalike}
  \renewcommand{\refname}{Bibliography}

  \bibliography{bibliography.bib}